\documentclass[10pt,journal,compsoc]{IEEEtran}
\usepackage{cite}
\usepackage{algorithmic}
\usepackage{array}
\usepackage{mdwmath}
\usepackage{mdwtab}
\usepackage{eqparbox}
\usepackage[tight,footnotesize]{subfigure}
\usepackage{cite}
\usepackage{url}
\usepackage{multicol,multirow}
\usepackage[cmex10]{amsmath}
\usepackage{makeidx}
\usepackage{diagbox}
\usepackage{overpic}
\usepackage{rotating,soul}
\usepackage{wrapfig}
\usepackage{booktabs}
\usepackage[usenames,dvipsnames]{color}
\usepackage{ragged2e}
\makeindex

\usepackage{geometry}
\DeclareGraphicsExtensions{.pdf,.jpg,.png}

\usepackage[pdfborder=false,colorlinks,bookmarksnumbered,bookmarksopen,linkcolor=black,citecolor=black,urlcolor=black]{hyperref}

\renewcommand{\tablename}{Table }

\newcommand{\figref}[1]{Fig.~\ref{#1}}

\newcommand{\equref}[1]{Eqn.~(\ref{#1})}

\def\ie{\emph{i.e.~}}
\def\eg{\emph{e.g.~}}
\def\etal{{\em et al.~}}
\def\etc{{\em etc.~}}

\graphicspath{{./Fig/}}

\begin{document}
	
\title{Deeply Supervised Salient Object Detection with Short Connections}

\author{Qibin Hou, Ming-Ming Cheng, Xiaowei Hu, Ali Borji, Zhuowen Tu,
  Philip~H.~S.~Torr
  \IEEEcompsocitemizethanks{
	\IEEEcompsocthanksitem Q. Hou, M.M. Cheng, and X. Hu are with
        CCCE, Nankai University.
        M.M. Cheng is the corresponding author (cmm@nankai.edu.cn).
	\IEEEcompsocthanksitem A. Borji is with the Center for Research
        in Computer Vision,
        University of Central Florida (aborji@crcv.ucf.edu)
	\IEEEcompsocthanksitem Z. Tuo is with the University of California
        at San Diego.
	\IEEEcompsocthanksitem P.H.S. Torr is with the University of Oxford.
	\IEEEcompsocthanksitem A preliminary version of this work appeared
            at CVPR \cite{hou2016deeply}.
			The source code are publicly available via our project page:
			\href{http://mmcheng.net/dss/}{http://mmcheng.net/dss/}.
	}
}

\markboth{IEEE TRANSACTIONS ON PATTERN ANALYSIS AND MACHINE INTELLIGENCE,~Vol.~xx,No.~xx,~xxx.~xxxx}%
{Hou \MakeLowercase{\textit{et al.}}: Deeply supervised salient object detection with short connections}
	
\IEEEcompsoctitleabstractindextext{%
\begin{abstract}
\justifying   
Recent progress on salient object detection is substantial,
benefiting mostly from the explosive development of Convolutional Neural Networks (CNNs).
Semantic segmentation and salient object detection algorithms developed
lately have been mostly based on Fully Convolutional Neural Networks (FCNs).
There is still a large room for improvement over the generic
FCN models that do not explicitly deal with the scale-space problem.
Holistically-Nested Edge Detector (HED) provides a skip-layer
structure with deep supervision for edge and boundary detection,
but the performance gain of HED on saliency detection is not obvious.
In this paper, we propose a new salient object detection method
by introducing short connections to the skip-layer structures
within the HED architecture.
Our framework takes full advantage of multi-level and multi-scale
features extracted from FCNs,
providing more advanced representations at each layer,
a property that is critically needed to perform segment detection.
Our method produces state-of-the-art results on $5$ widely
tested salient object detection benchmarks,
with advantages in terms of efficiency ($0.08$ seconds per image),
effectiveness, and simplicity over the existing algorithms.
Beyond that, we conduct an exhaustive analysis on
the role of training data on performance.
Our experimental results provide a more reasonable and
powerful training set for future research and fair comparisons.
\end{abstract}
		
}

\maketitle
\IEEEdisplaynontitleabstractindextext
\IEEEpeerreviewmaketitle
	
\IEEEraisesectionheading{\section{Introduction}\label{sec:Introduction}}
	
\IEEEPARstart{T}{he} goal in salient object detection is to identify the most
visually distinctive objects or regions in an image
and then segment them out from the background.
Different from other segmentation-like tasks,
such as semantic segmentation,
salient object detection pays more attention to
very few objects that are interesting and attractive.
Such a useful property allows salient object detection to commonly serve as the
first step to a variety of computer vision applications including image and
video compression \cite{guo2010novel,guo2017video}, 
image segmentation \cite{donoser2009saliency},
content-aware image editing \cite{cheng2010repfinder,ZhangC09},
object recognition \cite{rutishauser2004bottom},
weakly supervsied segmantic segmentation 
\cite{wei2016stc,HouDMWCT17,AdversErasingCVPR2017,wei2016learning}
visual tracking \cite{borji2012adaptive},
non-photo-realist rendering \cite{rosin2013artistic, han2013fast},
photo synthesis \cite{tog09Sketch2Photo,hu2013internet},
information discovery \cite{11tvc/Liu, zhu2012unsupervised},
image retrieval \cite{gao20123,cheng2017intelligent}, 
action recognition \cite{abdulmunem2016saliency} \etc

Earlier salient object detection methods were
mainly inspired by cognitive studies of
visual attention \cite{itti2001computational} 
where contrast plays the most important role in saliency detection.
Taking this fact into consideration,
various hand-crafted features have been designed,
employing either global or local cues
(See~\cite{borji2015salient,borji2013state} for reviews).
However, as these hand-crafted features are based on
the prior knowledge of existing datasets,
they cannot be extended to be successfully useful in all cases.
Although some works have attempted to develop different schemes to
combine these features rather than utilizing individual ones,
the resulting saliency maps are still far away
from being satisfactory, specially
when encountering complex and cluttered scenes.
To overcome the drawbacks caused by human priors,
learning based methods (\eg \cite{WangDRFI2017}) appear
to better integrate different types of features
to improve the generalization ability.
Nevertheless, because many fusion details are designed manually,
the enriched feature representations still suffer from
low contrast and fail to detect salient objects in
cluttered scenes.

\newcommand{\addFig}[1]{\includegraphics[width=0.114\linewidth]{#1}}
\begin{figure*}[t]
    \centering
    \begin{overpic}[width=\linewidth]{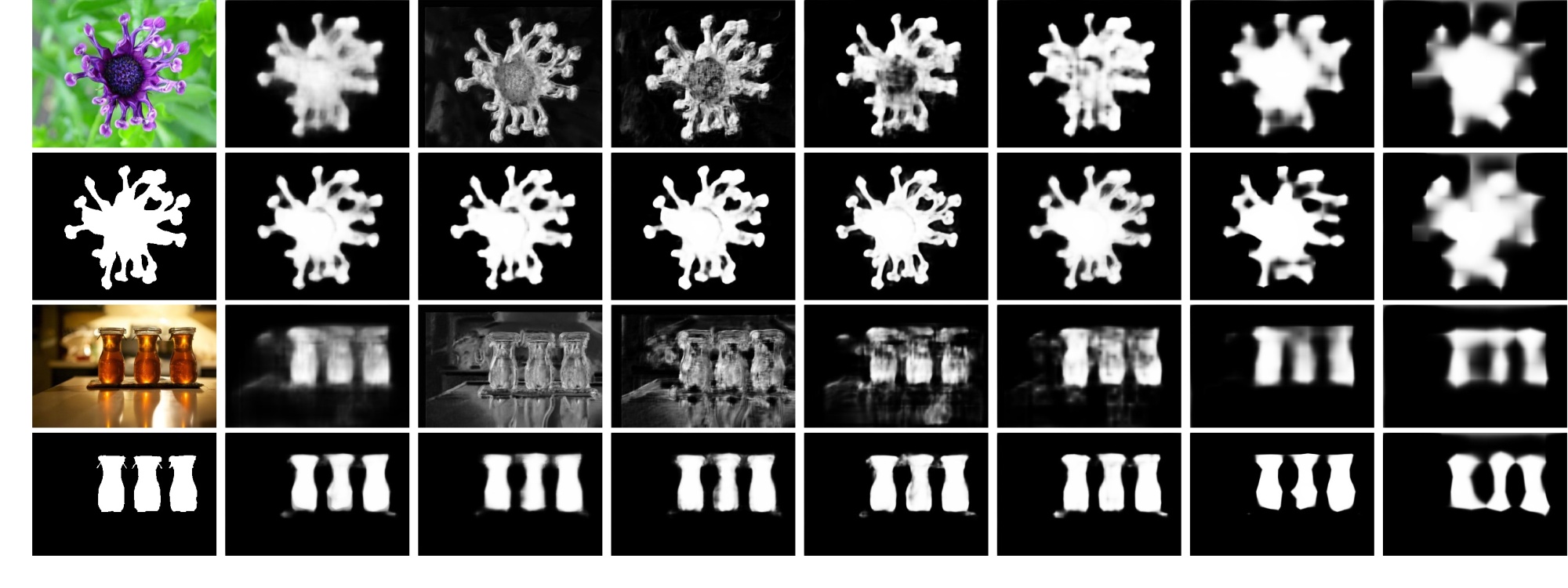}
        \put(1,0){(a) source \& GT}
        \put(16,0){(b) results}
        \put(29,0){(c) s-out 1}
        \put(41,0){(d) s-out 2}
        \put(53,0){(e) s-out 3}
        \put(65,0){(f) s-out 4}
        \put(78,0){(g) s-out 5}
        \put(90,0){(h) s-out 6}
        \put(0,2.5){\rotatebox{90}{\footnotesize Proposed}}
        \put(0,10){\rotatebox{90}{\footnotesize HED-based}}
        \put(0,19.5){\rotatebox{90}{\footnotesize Proposed}}
        \put(0,29){\rotatebox{90}{\footnotesize HED-based}}
    \end{overpic} \\
    \caption{Visual comparison of saliency maps produced by
    	the HED-based method \protect\cite{xie2017holistically} and ours.
    	Though saliency maps produced by deeper (4-6) side output (s-out)
        look similar, because of the introduced short connections,
        each shallower (1-3) side output can generate
        satisfactory saliency maps and hence a better output result.}
    \label{fig:comp_with_hed}
    \vspace{-10pt}
\end{figure*}

In a variety of computer vision tasks, such as
image classification \cite{krizhevsky2012imagenet,simonyan2014very},
semantic segmentation \cite{long2015fully},
edge detection \cite{xie2017holistically,RCFEdgeCVPR2017},
object detection \cite{li2017perceptual,girshick2015fast},
and pedestrian detection \cite{zhang2016faster},
convolutional neural networks (CNNs) \cite{lecun1998gradient}
have successfully broken the limits of traditional hand-crafted features.
%
%
The emergence of fully convolutional neural networks (FCNs) \cite{long2015fully}
have further boosted the development of these research areas,
providing a more principled learning method.
Such an end-to-end learning tool also motivates
recent research efforts of using FCNs for
salient object detection \cite{li2016deep, liu2016dhsnet}.
Benefiting from the enormous amount of parameters in FCNs,
a large margin of performance
gain has been made compared to previous approaches.
The holistically-nested edge detector (HED) \cite{xie2017holistically} model,
which explicitly deals with the scale space problem,
has led to large improvements over generic FCN models
in the context of edge detection.
Though the mechanism of fusing the multi-level features
extracted from different
scales provides a much more natural way to edge detection,
it is incompetent to do segmentation related tasks.
Edge detection is an easier task since it does not rely
too much on high-level semantic feature representations.
This explains why skip-layer structure with deep supervision in the HED model
does not lead to obvious performance gain for saliency detection.
Experimental results also support this statement
as shown in \figref{fig:comp_with_hed}.

In this paper, we focus on skip-layer structure with deep supervision.
Instead of simply fusing the multi-level features
extracted from different scales,
we consider such a problem in a top-down view.
As demonstrated in \figref{fig:comp_with_hed}, we observe that
1) deeper side outputs encode high-level semantic knowledge and
    hence can better locate where the salient objects are.
    However, due to the down-sampling operations in FCNs,
    the predicted maps are normally with irregular shapes
    especially when the input image is complex
    and cluttered (see the bottle image), and
2) shallower side outputs capture rich spatial information.
    They are capable of successfully highlighting the boundaries
    of those salient objects in spite of the resulting messy prediction maps.
%
Based on these phenomenons,
an intuitive idea for yielding better saliency maps
is to reasonably combine these multi-level features.
This motivates us to develop a new method for salient object detection
by introducing \emph{short connections} to the skip-layer structure
within the HED \cite{xie2017holistically} architecture.
By having a series of short connections
from deeper side outputs to the shallower ones,
our new framework offers two advantages:
\begin{enumerate}
\item high-level features can be transformed to shallower side-output layers
    and thus can help them better locate the most salient region, and
\item shallower side-output layers can learn rich low-level features that
    can help refine the sparse and irregular prediction maps from
    deeper side-output layers.
\end{enumerate}
By combining features from different levels, the resulting architecture
provides rich multi-scale feature maps at each layer,
a property that is essentially needed to do efficient salient object detection.
Our approach is fully convolutional and no other prior information such as
superpixels is needed.
It takes only 0.08s to produce a prediction map with resolution
of $300 \times 400$ pixels.
Other than improving the state-of-the-art results,
we conduct exhaustive analysis on the behavior of different training sets
as there is no universal training set for a fair comparison
in the salient object detection field.
Our goal is to offer a more unified training set and meanwhile
build a fair benchmarking environment for future research.

\begin{figure*}[t]
	\begin{center}
		\includegraphics[width=0.95\linewidth]{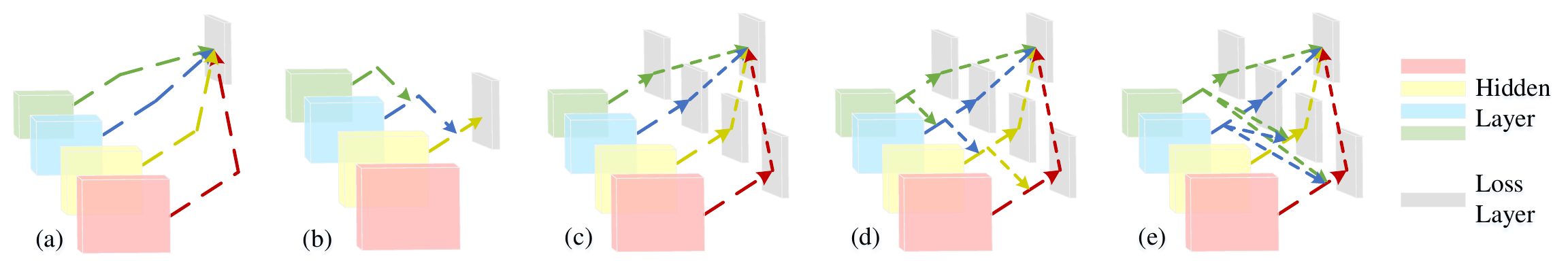}\\
		\caption{Illustration of different architectures.
        (a) Hypercolumn \protect\cite{hariharan2015hypercolumns},
        (b) FCN-8s \cite{long2015fully}
        (c) HED \protect\cite{xie2017holistically},
        (d) and (e) different patterns of our proposed architecture.
        As can be seen, a series of short connections are introduced
        in our architecture for combining the advantages of both deeper
        layers and shallower layers. More interestingly, the last one can be viewed as
        a generalized version of all the formers.
		} \label{fig:comp}
	\end{center}
    \vspace{-10pt}
\end{figure*}

\section{Related Works}

Over the past two decades,
an extremely rich set of saliency detection methods have been developed.
The majority of salient object detection methods are based on hand-crafted
local features \cite{itti1998model,xie2013bayesian,Qi2015}, 
global features
\cite{cheng2015global,ChengSaliency13iccv,liu2011learning}, 
or both \cite{borji2012exploiting,WangDRFI2017}.
A complete survey of these methods is beyond the scope of this paper and
we refer the readers to recent survey papers
\cite{borji2015salient,borji2014salient} for details.
Here, we mainly focus on discussing recent salient object detection methods based
on deep learning architectures.

\subsection{CNN-Based Saliency Models}

Compared with traditional methods that use hand-crafted features,
CNN-based methods have refreshed all the previous
state-of-the-art records in nearly every sub-field of computer vision,
including salient object detection.
In \cite{SuperCNN_IJCV2015}, He \etal presented a superpixel-wise
convolutional neural network architectures by utilizing hierarchical contrast features.
For each scale of superpixels, two contrast sequences were fed into convolutional
networks for building more advanced features.
Finally, different weights were learned to fuse
the multi-scale saliency maps together, yielding a much more confident one.
Li \etal \cite{li2015visual} proposed to use multi-scale features
extracted from a deep CNN to derive a saliency map.
By feeding different levels of image segmentation
into the deep CNN and aggregating multiple resulting features,
a stack of fully connected layers are then used to
determine on whether each segmented region is salient.
Wang \etal \cite{wang2015deep} predicted saliency maps by integrating
both local estimation and global search.
A deep neural network is first used to learn local patch features to provide each
pixel a saliency value.
Then, the local saliency map, global contrast, and geometric information are merged
together as the input to another deep neural network, which is used to predict the
saliency score of each region.
In \cite{zhao2015saliency}, Zhao \etal presented a multi-context
deep learning framework for salient object detection.
Two different CNNs are designed to independently
capture the global and local context
information of each segment patch.
A final regressor is used for final saliency decision of each segment patch.
Lee \etal \cite{lee2016deep} took into account both high-level semantic features
extracted from CNNs and hand-crafted features.
To combine them together, a unified fully connected neural network
was exploited to estimate saliency of each query region.
Liu \etal \cite{liu2016dhsnet} designed a two-stage deep network,
in which a coarse prediction map was produced,
followed by a recurrent CNN to refine the details of
the prediction map hierarchically and progressively.
In \cite{li2016deep}, a deep contrast network was proposed by leveraging the
contrast information of the input images.
It combined a pixel-level fully convolutional stream and
a segment-wise spatial pooling stream.
A fully connected conditional random field (CRF)
is also used for further refining
the prediction maps from the contrast network.
In \cite{wangsaliency}, Wang \etal proposed to leverage the advantages of recurrent
fully convolutional networks.
By doing so, their recurrent fully convolutional network
allowed them to continuously refine previous prediction maps
by correcting prediction errors.
A pre-training strategy using semantic segmentation data
is exploited for extracting
generic representations of salient objects.

\subsection{Skip-Layer Structures}

Very recently, great progress has been made in segment detection because of CNNs
and their flexible architectures.
Of these versatile structures,
skip-layer structures have been widely accepted by most researchers
owning to their capability of fusing multi-level and multi-scale features.
Early-stage skip-layer structures such as Hypercolumn \cite{hariharan2015hypercolumns}
and DCL \cite{li2016deep} have made breakthroughs in their respective fields.
They, however, only simply fuse the skip layers with
different scales for more advanced feature representation building
as shown in \figref{fig:comp}(a).
Differently, FCN-like structures \cite{long2015fully}
(see \figref{fig:comp}(b)) considered a better way
to utilize multi-level features,
gradually fusing the features from upper layers to lower ones.
In \cite{xie2017holistically}, Xie and Tu proposed a scheme
with deep supervision for each side output (skip layer).
Other than fusing all skip layers together,
a series of side losses are added after each side output
for preserving more details of the edge information.
\figref{fig:comp}(c) shows a simplified version of these architecture.

Despite the fact that multi-level and multi-scale features
have been taken into account and significant progress
has been made by these developments very recently,
there is still a large room for improvement over the generic CNN models
that do not explicitly deal with the scale-space problem.

\section{Deep Supervision with Short Connections} \label{section:architecture}

This section describes our approach and some implementation details.
Before that, let us first take a look at the observations.

\subsection{Observations}

As pointed out in most previous works, a good salient object
detection network should be deep enough such that multi-level
features can be learned.
Further, it should have multiple stages
with different strides so as to learn more inherent features from
different scales.
A good candidate for such requirements might be the HED network
\cite{xie2017holistically}, in which a series of side-output layers
are added after the last convolutional layer of each
stage in the VGGNet \cite{simonyan2014very}.
However, experimental results show that this
architecture is not suitable for salient object detection.
\figref{fig:comp_with_hed} provides such an illustration.
The reasons for this phenomenon are two-fold.
On the one hand, saliency detection, requiring homogeneous regions,
is quite different from edge detection that demands a special treatment.
A good saliency detection algorithm should be capable of extracting
the most visually distinctive objects and regions from an image
instead of simple edge information.
On the other hand, the features generated from lower stages
are too convoluted and the saliency maps obtained from the
deeper side-output layers are short of regularity.

To overcome the aforementioned problems, we propose a top-down method
to reasonably combine both low-level and high-level features for
accurate saliency detection.
The following subsections are dedicated to a detailed
description of the proposed approach.

\begin{figure}[tp!]
  \centering
  \includegraphics[width=\linewidth]{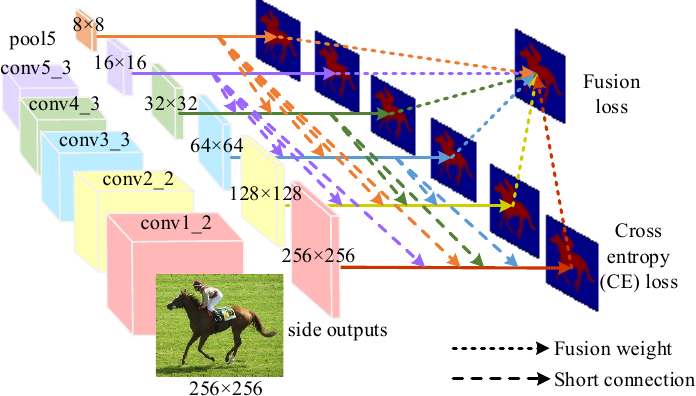}\\
  \caption{The proposed network architecture.
    Our architecture is based on VGGNet \protect\cite{simonyan2014very}
    for better comparison with previous CNN-based methods.
    As there are totally 6 different scales in VGGNet,
    6 side outputs are introduced, each of which is represented by different colors.
    Besides the side loss for each side output,
    a fusion loss is employed for capturing features of different levels.
  }\label{fig:arch}
  \vspace{-10pt}
\end{figure}

\subsection{HED-based saliency detection} \label{section:HED-based}

To better understand our proposed approach,
we start out with the standard HED architecture
\cite{xie2017holistically} as well as its
extended version, a special case of this work, for salient object
detection and gradually move on to our proposed architecture.

\subsubsection{HED architecture}

In the HED architecture \cite{xie2017holistically},
5 side outputs are introduced,
each of which is directly connected to
the last convolutional layer of each stage.
Let $T = \{(X_n,Z_n), n = 1,\ldots,N\}$ denote the training data set,
where $X_n=\{x_j^{(n)},j=1,\ldots,|X_n|\}$ is the input image and
$Z_n=\{z_j^{(n)},j=1,\ldots,|X_n|\}, z_j^{(n)} \in [0,1]$ denotes
the corresponding continuous ground truth saliency map for $X_n$.
In the sequel, we omit the subscript $n$ for notational convenience
since we assume the inputs are all independent of one another.
We denote the collection of all standard network layer parameters
as $\mathbf{W}$.
Without loss of generality, we further suppose that there are totally $M$ side outputs.
Each side output is associated with a classifier, in which
the corresponding weights can be represented by
$\mathbf{w} = (\mathbf{w}^{(1)},\mathbf{w}^{(2)},\ldots,\mathbf{w}^{(M)})$.
Thus, the side objective function of HED can be given by
\begin{equation}
L_{\text{side}}(\mathbf{W}, \mathbf{w}) =
\sum_{m=1}^{M}\alpha_{m} l_{\text{side}}^{(m)}\big(\mathbf{W}, \mathbf{w}^{(m)}\big),
\end{equation}
where $\alpha_{m}$ is the weight of the $m$th side loss and
$l_{\text{side}}^{(m)}$ denotes the image-level class-balanced cross-entropy
loss function \cite{xie2017holistically} for the $m$th side output.
Besides, a weighted-fusion layer is added to better capture the
advantage of each side output.
The fusion loss at the fusion layer can be expressed as
\begin{equation}
L_{\text{fuse}}(\mathbf{W}, \mathbf{w}, \mathbf{f}) = \sigma
\big(Z,h(\sum_{m=1}^{M}f_{m}A_{\text{side}}^{(m)})\big),
\end{equation}
where $\mathbf{f}=(f_1,\ldots,f_M)$ is the fusion weights,
$A_{\text{side}}^{(m)}$ are activations of the $m$th side output,
$h(\cdot)$ denotes the sigmoid function,
and $\sigma(\cdot,\cdot)$ denotes the distance between the ground truth map
and the fused predictions, which is set to be
image-level class-balanced cross-entropy loss
\cite{xie2017holistically}.
Therefore, the final loss function is given by
\begin{equation}
L_{\text{final}}\big(\mathbf{W}, \mathbf{w}, \mathbf{f}) =
L_{\text{fuse}}\big(\mathbf{W}, \mathbf{w}, \mathbf{f}) +
L_{\text{side}}\big(\mathbf{W}, \mathbf{w}).
\end{equation}

HED connects each side output to the last convolutional layer in each
stage of the VGGNet \cite{simonyan2014very}, respectively conv1\_2,
conv2\_2, conv3\_3, conv4\_3, conv5\_3.
Each side output is composed of a one-channel convolutional layer with the kernel size
$1\times1$ followed by an up-sampling layer for learning edge information.

\begin{figure}
	\centering
	\setlength\tabcolsep{7pt}
	\begin{tabular}{clccc} \specialrule{0.1em}{2pt}{2pt}
		No. & Layer & 1 & 2 & 3 \\ \specialrule{0.15em}{2pt}{2pt}
		1 & conv1\_2 & $128, 3 \times 3$ & $128, 3 \times 3$ & $1, 1 \times 1$ \\
		2 & conv2\_2 & $128, 3 \times 3$ & $128, 3 \times 3$ & $1, 1 \times 1$  \\
		3 & conv3\_3 & $256, 5 \times 5$ & $256, 5 \times 5$ & $1, 1 \times 1$  \\
		4 & conv4\_3 & $256, 5 \times 5$ & $256, 5 \times 5$ & $1, 1 \times 1$ \\
		5 & conv5\_3 & $512, 5 \times 5$ & $512, 5 \times 5$ & $1, 1 \times 1$ \\
        6 & pool5 & $512, 7 \times 7$ & $512, 7 \times 7$ & $1, 1 \times 1$ \\ \specialrule{0.1em}{2pt}{2pt}
	\end{tabular}
	\caption{Details of each side output. $(n, k \times k)$ means
		that the number of channels and the kernel size are $n$
		and $k$, respectively. ``Layer'' means which layer the
        corresponding side output is connected to. ``1"， ``2"，
        and ``3" represent three
		convolutional layers that are used in each side output.
		Note that the first two convolutional layers in each side
        output are followed by a ReLU layer for nonlinear transformation.}
    \label{tab:side_info}
\end{figure}

\subsubsection{Enhanced HED architecture} \label{src:enhanced_hed}

In this part, we extend the HED architecture for salient object detection.
During our experiments, we observe that deeper layers can
better locate the most salient regions,
so based on the architecture of HED we connect another side output
to the last pooling layer (pool5) in VGGNet \cite{simonyan2014very}.
Besides, since salient object detection is a more difficult task than edge detection,
we add two another convolutional layers with different filter channels
and spatial sizes in each side output, which can be found in \figref{tab:side_info}.
We use the same bilinear interpolation operation as in HED for up-sampling.
We also use a standard cross-entropy loss and compute the loss function
over all pixels in a training image $X = \{x_j, j=1,\ldots,|X|\}$ and
saliency map $Z = \{z_j, j=1,\ldots,|Z|\}$.
Our loss function can be defined as follows:
\begin{eqnarray*}
\hat{l}_{\text{side}}^{(m)}(\mathbf{W}, \hat{\mathbf{w}}^{(m)})
= -\sum_{z_j \in Z}z_j\log \text{Pr}\big(z_j = 1|X; \mathbf{W}, \hat{\mathbf{w}}^{(m)}\big)
\end{eqnarray*}
\begin{equation}\label{eqn:SCE}
+~(1 - z_j)\log \text{Pr}\big(z_j = 0|X; \mathbf{W}, \hat{\mathbf{w}}^{(m)}\big),
\end{equation}
where $\text{Pr}\big(z_j = 1|X; \mathbf{W}, \hat{\mathbf{w}}^{(m)}\big)$
represents the probability of the activation value at location
$j$ in the $m$th side output,
which can be computed by $h(a_j^{(m)})$, where
$\hat{A}_{\text{side}}^{(m)} = \{a_j^{(m)}, j=1,\dots,|X|\}$
are activations of the $m$th side output.
Similar to \cite{xie2017holistically}, we add a
weighted-fusion layer to connect each side activation.
The loss function at the fusion layer in our case can be represented by
\begin{equation}
	\hat{L}_{\text{fuse}}(\mathbf{W}, \hat{\mathbf{w}}, \mathbf{f}) =
	\hat{\sigma}\big(
    \begin{matrix}
    Z, \sum_{m=1}^{\hat{M}}f_{m}\hat{A}_{\text{side}}^{(m)}
    \end{matrix}
    \big),
\end{equation}
where $\hat{A}_{\text{side}}^{(m)}$ is the new activations
of the $m$th side output\footnote{We add a new side output in our enhanced HED architecture.}, $\hat{M} = M + 1$,
and $\hat{\sigma}(\cdot,\cdot)$ represents the distance between
the ground truth map and the new fused predictions,
which has the same form as in \equref{eqn:SCE}.

A comparison of salient object detection results between the
original HED and enhanced HED is shown in \figref{fig:patterns}.
It can be easily found that a large margin of about $3\%$ improvement has been achieved.
In spite of such improvement, as shown in \figref{fig:comp_with_hed},
the saliency maps from shallower side outputs still look messy and
the deeper side outputs produce irregular results as well.
In addition, the deeper side outputs can indeed locate the salient objects,
but some detailed information is still lost.

\subsection{Short connections}
\label{section:short}

\begin{figure}[tpb]
	\begin{center}
		\includegraphics[width=0.9\linewidth]{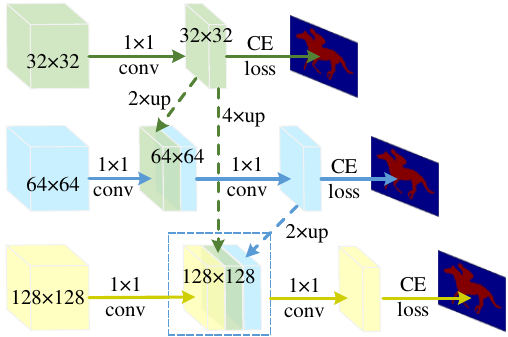}\\ 
		\caption{Illustration of short connections in \protect\figref{fig:arch}.
		}\label{fig:sConnect}
	\end{center}
    \vspace{-10pt}
\end{figure}

The insight of our approach is that deeper side outputs are capable of
finding the location of salient regions but at the expense of the loss of details,
while shallower ones focus on low-level features but are short of global information.
These phenomenons inspire us to utilize the following way to
appropriately combine different side outputs such that the
most visually distinctive objects can be extracted.

\subsubsection{Formulation}

Mathematically, our new side activations $\tilde{R}_{\text{side}}^{(m)}$
at the $m$th side output can be given by
\begin{equation}
\begin{aligned}
\tilde{R}_{\text{side}}^{(m)} =
\begin{cases}
\begin{matrix}
    \sum_{i=m+1}^{\hat{M}}r_{i}^{m}\tilde{R}_{\text{side}}^{(i)}
    \end{matrix} + \hat{A}_{\text{side}}^{(m)}, \text{~for $m = 1, \ldots , 5$} \\
\hat{A}_{\text{side}}^{(m)}, \text{~for $m = 6$}  \\
\end{cases}
\end{aligned}
\end{equation}
where $r_{i}^{m}$ is the weight of short connection from side output $i$
to side output $m$ ($i > m$).
We can drop out some short connections by directly setting $r_{i}^{m}$
to 0.
The new side loss function and fusion loss function can be respectively
represented by
\begin{equation}
\tilde{L}_{\text{side}}(\mathbf{W}, \tilde{\mathbf{w}}, \mathbf{r}) = \sum_{m=1}^{\hat{M}}\alpha_{m} \tilde{l}_{\text{side}}^{(m)}\big(\mathbf{W}, \tilde{\mathbf{w}}^{(m)}, \mathbf{r}\big)
\end{equation}
and
\begin{equation} \label{eqn:fuse}
	\tilde{L}_{\text{fuse}}(\mathbf{W}, \tilde{\mathbf{w}}, \mathbf{f}, \mathbf{r}) =
	\hat{\sigma}\big(
    \begin{matrix}
    Z, \sum_{m=1}^{M}f_{m}\tilde{R}_{\text{side}}^{(m)}
    \end{matrix}
    \big),
\end{equation}
where $\mathbf{r} = \{r_{i}^{m}\}, i>m$.
Note that this time $\tilde{l}_{\text{side}}^{(m)}$ represents the
standard cross-entropy loss which we have defined in \equref{eqn:SCE}.
Thus, our new final loss function can be written as
\begin{equation}
\tilde{L}_{\text{final}}\big(\mathbf{W}, \tilde{\mathbf{w}}, \mathbf{f}, \mathbf{r}) =
\tilde{L}_{\text{fuse}}\big(\mathbf{W}, \tilde{\mathbf{w}}, \mathbf{f}, \mathbf{r}) +
\tilde{L}_{\text{side}}\big(\mathbf{W}, \tilde{\mathbf{w}}, \mathbf{r}).
\end{equation}

\subsubsection{Construction}

The backbone of our new architecture is the enhanced HED which has been described in
Section~\ref{src:enhanced_hed}.
\figref{fig:sConnect} illustrates how to construct short connections from side output 4 to side output 2.
The score maps in side outputs 3 and 4 are first upsampled by simple bilinear interpolation
and then concatenated to the original score map in side output 2.
The hyper-parameters of bilinear interpolation can be derived according to the context.
As salient object detection is a class-agnostic task, we further weight the foregoing
score maps which have been enclosed by a dashed bounding box in \figref{fig:sConnect}
and introduce another $1\times1$ convolutional layer as the new score map of side output 2.
A similar approach can be used for side outputs to which more than one short connection is connected.
For instance, let us assume that 3 short connections are connected to side output 2.
There would be 4 score maps being concatenated together within the dashed bounding box.

Our architecture can be functionally considered as two closely connected stages, which we call
\textit{saliency locating stage} and \textit{details refinement stage}, respectively.
The main focus of saliency locating stage is on looking for
the most salient regions in a given image.
For details refinement stage, we introduce a top-down method, a series of
short connections from deeper side-output layers to shallower
ones.
The reason for such a consideration is that with the help of
deeper side information, lower side outputs can both accurately predict the
salient objects and refine the results from deeper side outputs,
resulting in dense and accurate saliency maps.
We further test the effectiveness of our proposed architecture by
running a number of ablation experiments and
showing the corresponding quantitative and visual results in the next section.

\subsection{Implementation Details}

Our network is based on the publicly available Caffe library
\cite{jia2014caffe} and the open implementation of
FCN \cite{long2015fully}.
As mentioned above, we choose VGGNet \cite{simonyan2014very} as our
pre-trained model for better comparison with other works.
%

\subsubsection{Inference}

Although a series of short connections are introduced,
the quality of the prediction maps produced by the deeper
and the shallower side outputs is still unsatisfactory.
Regarding this fact, during the testing phase,
we adopt a more complicated combination of these side outputs.
Let $\tilde{Z}_{1}, \cdots, \tilde{Z}_{6}$ denote the score map of each side output, respectively.
They can be computed by $\tilde{Z}_{m} = h(\tilde{R}_{\text{side}}^{(m)})$.
Recall that $h(\cdot)$ in our case is the sigmoid function.
Therefore, the fusion output map can be computed by
\begin{equation}
	\tilde{Z}_{\text{fuse}} = h\big(\sum_{m=2}^{4}f_{m}\tilde{R}_{\text{side}}^{(m)}\big).
\end{equation}
To avoid the negative effect caused by the bad quality of the prediction map
from the deepest and shallowest side outputs,
we also use $\tilde{Z}_{2}, \hat{Z}_{3}$, and $ \hat{Z}_{4}$
to help further fill in the lost details.
As a result, the final output map during inference can be represented by
\begin{equation} \label{eqn:final_output}
	\tilde{Z}_{\text{final}} = \text{Mean}(\tilde{Z}_{\text{fuse}}, \tilde{Z}_{2}, \tilde{Z}_{3}, \tilde{Z}_{4}).
\end{equation}
Surprisingly, we found that such a combination do help improve the results by a little margin.
This is due to the fact that although the fusion output map incorporates the aggregation of each side output,
some detailed information in the fusion output map is still missed.
Regarding the quality of each side output map (see \figref{fig:comp_with_hed}), we decide to use
\equref{eqn:final_output} as the final output map.

\subsubsection{Smoothing Method}

Though our model can precisely find the salient objects in an image,
the boundary information of the resulting saliency maps is still lost for those complex scenes.
To further improve spatial coherence and quality of our saliency
maps, we adopt the fully connected conditional random field
(CRF) method \cite{krahenbuhl2012efficient} as a selective
layer during the inference phase.

The energy function of CRF is given by
\begin{equation}
E(\mathbf{x}) = \sum_{i}\theta_{i}(x_{i}) + \sum_{i,j}\theta_{ij}(x_{i}, x_{j}),
\end{equation}
where $\mathbf{x}$ is the label prediction for pixels.
To make our model more competitive, instead of directly using
the predicted maps as the input of the unary term, we leverage the
following unary term
\begin{equation} \label{eqn:unary_term}
\theta_{i}(x_{i}) = -\frac{\log \hat{S}_{i}}{\tau h(\hat{S}_{i})},
\end{equation}
where $\hat{S}_{i}$ denotes normalized saliency value of pixel
$x_{i}$, $h(\cdot)$ is the sigmoid function, and $\tau$ is a scale parameter.
The pairwise potential is defined as
\begin{equation}
\begin{aligned}
\theta_{ij}(x_{i}, x_{j}) = \mu(x_{i}, x_{j})\biggl[w_{1}\exp\biggl(-\frac{\lVert p_{i} - p_{j} \rVert^{2}}{2\sigma_{\alpha}^{2}} - \\
\frac{\lVert I_{i} - I_{j} \rVert^{2}}{2\sigma_{\beta}^{2}}\biggr) + w_{2}\exp\biggl(-\frac{\lVert p_{i} - p_{j}\rVert^{2}}{2\sigma_{\gamma}^{2}}\biggr)\biggr],
\end{aligned}
\end{equation}
where $\mu(x_{i}, x_{j}) = 1$ if $x_{i} \ne x_{j}$ and
zero, otherwise. $I_{i}$ and $p_{i}$ are pixel value and
position of $x_{i}$, respectively. Parameters
$w_{1}, w_{2}, \sigma_{\alpha}, \sigma_{\beta},$ and $\sigma_{\gamma}$
control the importance of each Gaussian kernel.

In this paper, we employ a publicly available implementation of
\cite{krahenbuhl2012efficient}, called PyDenseCRF
\footnote{https://github.com/lucasb-eyer/pydensecrf}.
Since there are only two classes in our case, we use the
inferred posterior probability of each pixel being salient
as the final saliency map directly.

\subsubsection{Parameters}
The hyper-parameters used in this work include learning rate (1e-8),
weight decay (0.0005), momentum (0.9), loss weight for each side output (1).
We use full-resolution images to train our network, and
the mini-batch size is set to 10.
The kernel weights in newly added convolutional layers are all
initialized with random numbers.
Our fusion layer weights are all initialized with 0.1667
in the training phase.
The parameters in the fully connected CRF are determined
using cross validation on the validation set.
In our experiments, $\tau$ is set to 1.05, and $w_{1}, w_{2}$,
$\sigma_{\alpha}, \sigma_{\beta},$ and $\sigma_{\gamma}$ are
set to 3.0, 3.0, 60.0, 8.0, and 5.0, respectively.

\section{Experiments and Analyses}

In this section, we introduce utilized datasets and evaluation
criteria and report the performance of our proposed approach.
Besides, a number of ablation experiments are performed for
analyzing the importance of each component of our approach.

\subsection{Datasets}

We evaluate our approach on
5 representative datasets, including MSRA-B \cite{liu2011learning},
ECSSD \cite{yan2013hierarchical}, HKU-IS \cite{li2015visual},
PASCALS \cite{li2014secrets},
and SOD \cite{martin2001database, movahedi2010design}, all of
which are available online.
These datasets all contain a large number of images as well as
well-segmented annotations and have been widely used recently.

MSRA-B contains 5,000 images from hundreds of different categories.
Because of its diversity and large quantity, MSRA-B has been
one of the most widely used datasets in salient object detection literature.
Most images in this dataset have only one salient object,
and hence it has gradually become a standard dataset
for evaluating the capability of processing simple scenes.
ECSSD contains 1,000 semantically meaningful but structurally complex natural
images.
HKU-IS is another large-scale dataset that contains more than
4000 challenging images.
Most of images in this dataset have low contrast with more than one salient object.
PASCALS contains 850 challenging images (each composed of several objects), all of which are chosen
from the validation set of the PASCAL VOC 2010 segmentation dataset.
We also evaluate our system on the SOD dataset, which is a subset of  the BSDS dataset.
It contains 300 images, most of which possess multiple salient objects.
All of these datasets consist of ground truth human annotations.

In order to preserve the integrity of the evaluation and obtain
a fair comparison with existing approaches, we utilize the same
training and validation sets
as in \cite{WangDRFI2017} and
test over all of the datasets using the same model.

\subsection{Evaluation Metrics}

We use three universally-agreed, standard metrics
(see also \cite{borji2015salient,cheng2015global,perazzi2012saliency,borji2015salient})
to evaluate our model including precision-recall curves,
F-measure, and the mean absolute
error (MAE).
For a given continuous saliency map $S$,
we convert it to a binary mask $B$ using a threshold.
Then its precision and recall are computed as
$precision = {|B \cap Z|}/{|B|}$ and $recall={|B \cap Z|}/{|Z|},$
respectively, where $|\cdot|$ accumulates the non-zero entries in a mask.
Averaging the precision and recall values over the saliency maps
of a given dataset yields the PR curve.

To comprehensively evaluate the quality of a saliency map, the
F-measure metric is used, which is defined as
\begin{equation}
F_{\beta} = \frac{(1 + \beta^2)Precision \times Recall}{\beta^2 Precision + Recall}.
\end{equation}
As suggested by previous works,  we choose $\beta^2$ to be 0.3
for stressing the importance of the precision value.

Let $\hat{S}$ and $\hat{Z}$ denote the continuous saliency map
and the ground truth that are normalized to $[0,1]$.
The mean absolute error (MAE) score can be computed as
\begin{equation}
MAE=\frac{1}{H \times W}\sum_{i=1}^{H}\sum_{j=1}^{W}|\hat{S}(i,j)=\hat{Z}(i,j)|.
\end{equation}

\subsection{Ablation Analysis}

\begin{figure*}
	\centering
    \small
	\setlength\tabcolsep{5.0pt}
	\begin{tabular}{cccccccc} \toprule[1pt]
		\multirow{1}*{}
		No.  & Side output 1 & Side output 2 & Side output 3 & Side output 4 & Side output 5 & Side output 6 & $F_\beta$ \\ \midrule[1pt]
        1 &	$(128, 3 \times 3) \times 2$ & $(128, 3 \times 3) \times 2$ & $(256, 5 \times 5) \times 2$ & $(512, 5 \times 5) \times 2$ & $(1024, 5 \times 5) \times 2$ & $(1024, 7 \times 7) \times 2$ & \textbf{0.830}  \\
        2 & $(128, 3 \times 3) \times 1$ & $(128, 3 \times 3) \times 1$ & $(256, 5 \times 5) \times 1$ & $(256, 5 \times 5) \times 1$ & $(512, 5 \times 5) \times 1$ & $(512, 7 \times 7) \times 1$ & 0.815  \\
        3 & $(128, 3 \times 3) \times 2$ & $(128, 3 \times 3) \times 2$ & $(256, 3 \times 3) \times 2$ & $(256, 3 \times 3) \times 2$ & $(512, 5 \times 5) \times 2$ & $(512, 5 \times 5) \times 2$ & 0.820  \\
        4 & $(128, 3 \times 3) \times 2$ & $(128, 3 \times 3) \times 2$ & $(256, 5 \times 5) \times 2$ & $(256, 5 \times 5) \times 2$ & $(512, 5 \times 5) \times 2$ & $(512, 7 \times 7) \times 2$ & \textbf{0.830}  \\ \bottomrule[1pt]
	\end{tabular}
    \caption{Comparisons of different side output settings and their performance on
    PASCALS dataset \protect\cite{li2014secrets}. $(c, k \times k) \times n$ means that
    there are $n$ convolutional layers with $c$ channels and size $k \times k$.
    Note that the last convolutional layer in each side output is unchanged as listed in
    \figref{tab:side_info}. In each setting, we only modify one parameter while keeping all
    others unchanged so as to emphasize the importance of each chosen parameter.}
	\label{fig:side_settings}
\end{figure*}

We experiment with different design options and different short
connection patterns to illustrate the
effectiveness of each component of our method.

\subsubsection{Various Short Connection Patterns}
Our architecture as shown in \figref{fig:arch} is
so flexible that can be regarded as the generalized model
of most existing architectures,
such as those depicted in \figref{fig:comp}.
To better show the strength of our proposed approach,
we use different network architectures as listed
in \figref{fig:comp} for salient object detection.
Besides the Hypercolumns architecture \cite{hariharan2015hypercolumns}
and the HED-based architecture \cite{xie2017holistically},
we implement three representative patterns using our proposed approach.
The first one is formulated as follows, which is a similar
architecture to \figref{fig:comp}(d).
\begin{equation} \label{eqn:pattern1}
\begin{aligned}
\tilde{R}_{\text{side}}^{(m)} =
\begin{cases}
r_{m+1}^{m}\tilde{R}_{\text{side}}^{(m+1)} + \hat{A}_{\text{side}}^{(m)}, \text{~for $m = 1,\ldots,5$} \\
\hat{A}_{\text{side}}^{(m)}. \text{~for $m = 6$}  \\
\end{cases}
\end{aligned}
\end{equation}
The second pattern is represented as follows which is much more
complex than the first one.
\begin{equation} \label{eqn:pattern2}
\begin{aligned}
\tilde{R}_{\text{side}}^{(m)} =
\begin{cases}
\begin{matrix}
\sum_{i = m+1}^{m+2}r_i^m\tilde{R}_{\text{side}}^{(i)} +
\hat{A}_{\text{side}}^{(m)}, \text{~for $m = 1, 2, 3, 4$}
\end{matrix} \\
\hat{A}_{\text{side}}^{(m)}. \text{~for $m = 5, 6$}  \\
\end{cases}
\end{aligned}
\end{equation}
The last pattern, the one used in this paper, is given by
\begin{equation} \label{eqn:pattern3}
\begin{aligned}
\tilde{R}_{\text{side}}^{(m)} =
\begin{cases}
\begin{matrix}
\sum_{i = 3}^{6}r_i^m\tilde{R}_{\text{side}}^{(i)} +
\hat{A}_{\text{side}}^{(m)}, \text{~for $m = 1, 2$}
\end{matrix} \\
r_{5}^{m}\tilde{R}_{\text{side}}^{(5)} + r_{6}^{m}\tilde{R}_{\text{side}}^{(6)} +
\hat{A}_{\text{side}}^{(m)}, \text{~for $m = 3, 4$}  \\
\hat{A}_{\text{side}}^{(m)}. \text{~for $m = 5, 6$}  \\
\end{cases}
\end{aligned}
\end{equation}
The quantitative results are listed in \figref{fig:patterns}.
As can be seen from \figref{fig:patterns},
by adding another side output and two
additional convolutional layers in each side output,
we have a performance gain of 2.5 points in terms of F-measure.
In addition, with the increase of short connections,
our approach gradually achieves better performance.
Although there is no performance gain obtained
when Pattern 1 is used compared with the enhanced HED structure,
a gain of 0.8 points can be achieved when we turn to Pattern 2.
Another 0.6 points gain can also be obtained when Pattern 3 is considered.

\subsubsection{Details of Side-Output Layers}
We run several ablation experiments to explore the best side output settings.
The detailed information of each side-output layer
in each experiment has been shown in \figref{fig:side_settings}.
We use Pattern 3 in \figref{fig:patterns} as our baseline model.
To highlight the importance of different parameters, we adopt
the variable-controlling method that only changes one parameter
at a time.
Besides, all the results are tested on PASCALS dataset for fair comparison.
Compared with the fourth experiment,
the first one exploits more channels but the same F-measure score is obtained.
This means that more channels for each side output
cannot bring in additional performance gain.
In the second experiment,
we tried to reduce 1 convolutional layer in each side output
but it turns out that such an operation decreases the performance by 1.5 points.
In spite of a small decrease,
it is enough to account for the importance of
introducing two convolutional layers in each side output.
Furthermore, we attempt to reduce the large kernel size
in deeper side outputs.
Similarly, this leads to a slight decrease in F-measure.
All the above experiments demonstrate that
the side output settings we use are reasonable and appropriate.

\begin{figure}
	\centering
    \small
	\setlength\tabcolsep{13pt}
	\begin{tabular}{ccc} \toprule[1pt]
		Scheme & Architecture & F-measure \\ \midrule[1pt]
		1 & Hypercolumns \cite{hariharan2015hypercolumns} &	0.818 \\
		2 & Original HED \cite{xie2017holistically} & 0.791  \\
        3 & Enhanced HED & 0.816  \\
		4 & Pattern 1 (\equref{eqn:pattern1}) & 0.816  \\
        5 & Pattern 2 (\equref{eqn:pattern2}) & 0.824   \\
		6 & $\text{Pattern 3}^{*}$ (\equref{eqn:pattern3}) & \textbf{0.830}  \\ \bottomrule[1pt]
	\end{tabular}
	\caption{The performance of different architectures on PASCALS dataset
    \protect\cite{li2014secrets}. '*' represents the pattern used in this paper.}
    \label{fig:patterns}
    \vspace{-10pt}
\end{figure}

\subsubsection{Upsampling Operation}
In our approach, we use the in-network bilinear interpolation to
perform upsampling in each side output.
As implemented in \cite{long2015fully}, we use fixed deconvolutional
kernels for our side outputs with different strides.
Since the prediction maps generated by deep side-output layers are not
dense enough, we also try to use the ``hole algorithm'' to make the
prediction maps in deep side outputs denser.
We adopt the same technique as in \cite{li2016deep}.
However, according to our experiments, using such a method yields a worse
performance.
We notice that as the fusion prediction map gets denser, some non-salient
pixels are wrongly predicted as salient ones even though
the CRF is used thereafter.
The F-measure score on the validation set is decreased by nearly
1\%.

\subsubsection{Data Augmentation}
Data augmentation has been
proven to be very useful in many learning-based vision tasks.
As done in most previous works, we flip all the training images horizontally,
resulting in an augmented image set with twice larger than the original one.
We found that such an operation further improves the performance by more than 0.5\%.
In addition, we also try to crop the input images to a fixed size $321\times321$.
However, experimental results show that
such an operation decrease our performance by more than 0.5 points.
This may be because input images with full size contain richer information
that allows our network to better capture the salient objects.

\subsubsection{Different Backbones}
We also extend our work by replacing the VGGNet with
ResNet-101 \cite{He2016} as the backbone.
Taking into account the network structure of ResNet-101,
we only use the bottom 5 side outputs in \figref{tab:side_info},
which are connected to conv1, res2c, res3b3, res4b22, and res5c, respectively.
We keep other settings unchanged.
We show the results on the bottom of \figref{tab:comparision}.
With the same training set,
there is a further one-point improvement on each dataset
in terms of F-measure score on average.

\subsubsection{The Proposed CRF Model}
Most previous works \cite{krahenbuhl2012efficient,li2016deep}
only use the negative log likelihood as the unary term in their CRF model.
Differently from them,
we introduce a modulating factor that aims to
give positive predictions more confidence
as shown in \equref{eqn:unary_term}.
This is reasonable as most of the predictions
are correct through observing the MAE scores.
In our experiments, we found that adding such a modulating factor
helps little on improving the F-measure scores but
is able to further reduce the MAE scores
(\ie, reduce wrong predictions) by around 0.3 points.

\subsection{Comparison with the State-of-the-art}

We compare the proposed approach with 7 recent CNN-based methods,
including MDF \cite{li2015visual}, DS \cite{li2016deepsaliency}, DCL
\cite{li2016deep}, ELD \cite{lee2016deep}, MC \cite{zhao2015saliency},
RFCN \cite{wangsaliency}, and DHS \cite{liu2016dhsnet}.
Four classical methods are also considered including RC \cite{cheng2015global},
CHM \cite{li2013contextual}, DSR \cite{li2013saliency},
and DRFI \cite{WangDRFI2017}, which have been
proven to be the best in the benchmark study of Borji
\etal~\cite{borji2015salient}.
It is worth mentioning that though more training images
is able to bring us better results as shown in
\figref{tab:dataset_comparison},
our results here are mainly based on 2500 training images
from MSRA-B dataset for fair comparison with existing works.

\renewcommand{\addFig}[1]{\includegraphics[width=0.079\linewidth]{Example/#1}}
\newcommand{\addFigs}[1]{\addFig{#1.jpg} & \addFig{#1.png} & \addFig{#1_dss.png}
   & \addFig{#1_DCL.png} & \addFig{#1_dhsnet.png} & \addFig{#1_rfcn.png} &
   \addFig{#1_ds.png} & \addFig{#1_MDF.png} & \addFig{#1_ELD.png} &
   \addFig{#1_mc.png} & \addFig{#1_DRFI.png} & \addFig{#1_DSR.png} \\
}

\newcommand{\figDescr}[1]{\specialrule{0.05em}{0pt}{5pt} \multicolumn{12}{l}{#1}}

\begin{figure*}[tp]
	\centering
	\setlength\tabcolsep{1pt}
	\begin{tabular}{cccccccccccc}
		\multicolumn{12}{l}{Simple Scene $|$ Complex Scene $|$ Center Bias} \\
        \addFigs{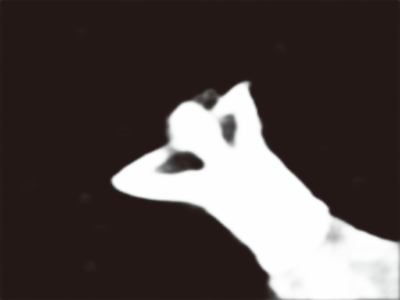}
        \addFigs{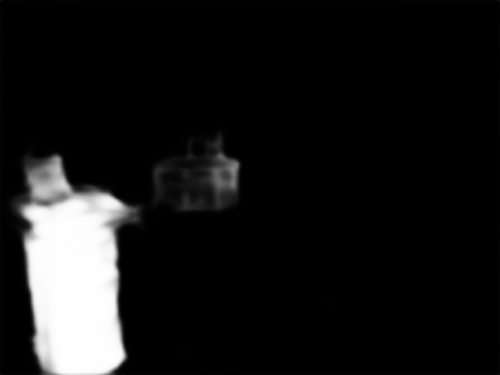}
		\figDescr{Complex Scene $|$ Small Object $|$ Low Contrast} \\
        \addFigs{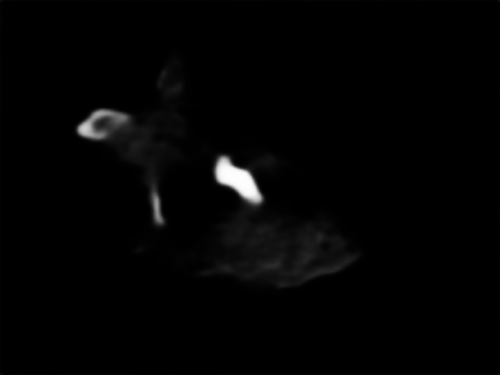}
        \addFigs{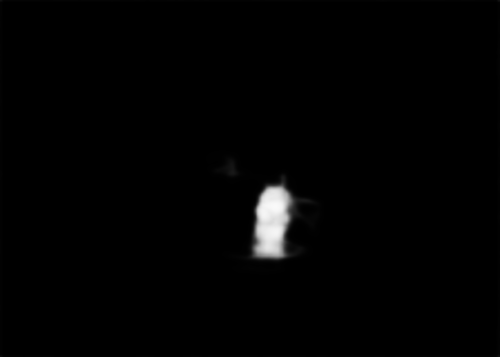}
        \figDescr{Low Contrast $|$ Complex Texture} \\
        \addFigs{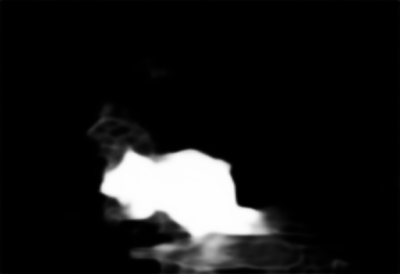}
        \addFigs{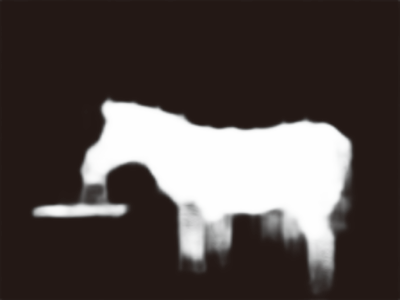}
        \figDescr{Large Object $|$ Low Contrast} \\
        \addFigs{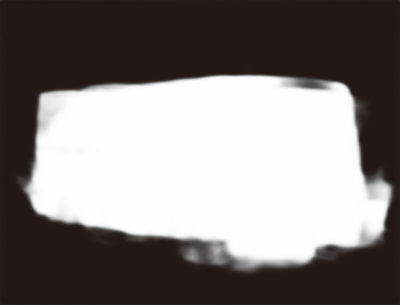}
        \addFigs{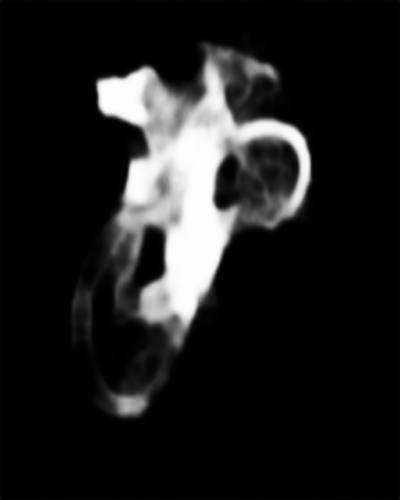}
        \figDescr{Multiple Object $|$ Large Object $|$ Complex Scene} \\
        \addFigs{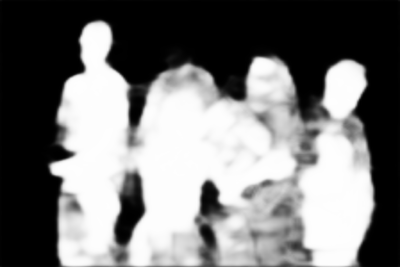}
        \addFigs{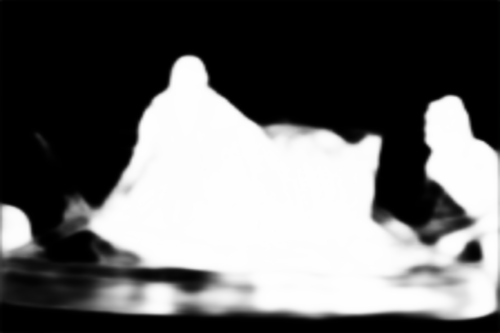}
        \figDescr{Multiple Object $|$ Transparent Object} \\
        \addFigs{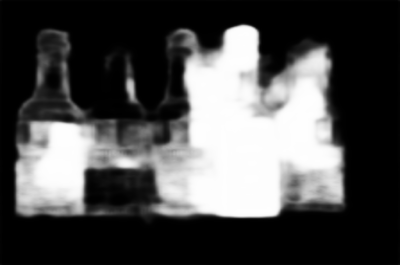}
		Source  & GT  & Ours  & DCL & DHS & RFCN & DS & MDF & ELD & MC  & DRFI & DSR
	\end{tabular}
	\caption{Selected results from various datasets.
      We split the selected images into multiple groups,
      which are separated by solid lines.
      To better show the capability of processing different
      scenes for each approach,
      we highlight the features of images in each group.
    }\label{fig:examples}
    \vspace{-10pt}
\end{figure*}

\renewcommand{\addFig}[1]{\begin{overpic}[width=0.32\linewidth]{#1-pr.pdf}
        \put(21,8.0){Ours}
        \put(21,12.6){DCL \cite{li2016deep}}
        \put(21,17.2){DHS \cite{liu2016dhsnet}}
        \put(21,21.8){RFCN \cite{wangsaliency}}
        \put(21,26.4){MDF \cite{li2015visual}}
        \put(21,31){ELD \cite{lee2016deep}}
        \put(21,35.6){MC \cite{zhao2015saliency}}
        \put(21,40.2){DRFI \cite{WangDRFI2017}}
        \put(21,44.8){DSR \cite{li2013saliency}}
        \put(21,49.4){CHM \cite{li2013contextual}}
        \put(21,54){RC \cite{cheng2015global}}
    \end{overpic}}

\begin{figure*}[t]
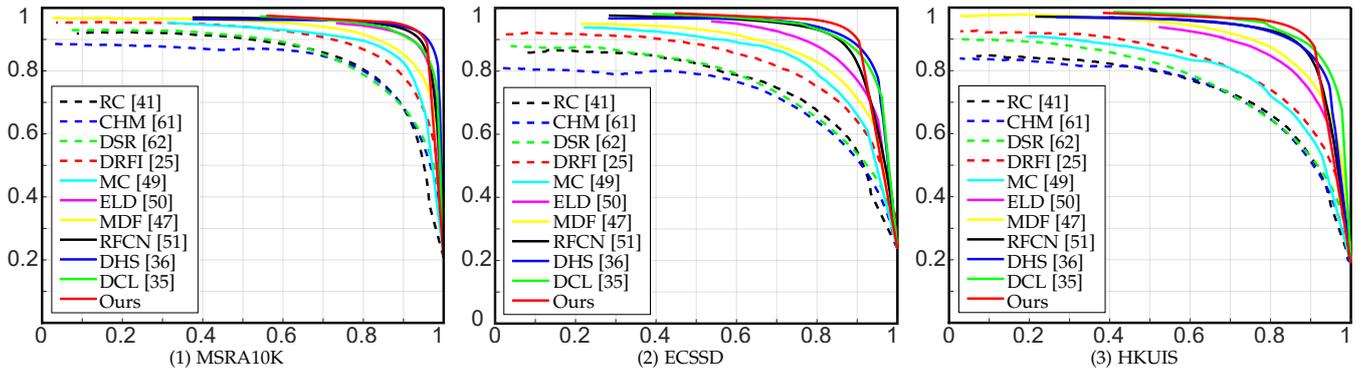

    \centering
    \scriptsize
	\setlength\tabcolsep{3pt}
    \begin{tabular*}{\textwidth}{ccc}
        \addFig{msra} & \addFig{ecssd} & \addFig{hkuis} \\
        (1) MSRA10K  &  (2) ECSSD & (3) HKUIS \\
    \end{tabular*}
    \caption{Precision (vertical axis) recall (horizontal axis)
        curves on three popular salient object  datasets.
    }\label{fig:pr_curve}
\end{figure*}

\subsubsection{Visual Comparison}
To exhibit the superiority of our proposed approach
compared against the above-mentioned methods,
we select multiple representative images from different
datasets which incorporate a variety of difficult circumstances,
including complex scenes, salient objects with center bias,
salient objects with different sizes,
low contrast between foreground and background, etc.,
and show the visual comparisons in \figref{fig:examples}.
We manually split the selected images into
multiple groups which are separated by solid lines.
We also give each group multiple tags
describing their properties.

Taking all circumstances into account,
it can be easily seen that our proposed method
not only highlights the right salient regions
but also produces coherent boundaries.
It is also worth mentioning that thanks to the short connections,
our approach gives salient regions more confidence,
yielding higher contrast
between salient objects and the background.
More importantly, it generates connected regions,
which greatly strengthens the ability of our model.
These advantages make our results very close to
the ground truth and hence better than other methods
in almost all circumstances which are shown in \figref{fig:examples}.

\newcommand{\first}[1]{{\bf #1}}
\newcommand{\second}[1]{\underline{#1}}
\newcommand{\TheMeasures}{$F_\beta$ & {\sc mae}}

\begin{figure*}
    \centering
    \small
    \setlength\tabcolsep{6.8pt}
    \begin{tabular*}{\textwidth}{lcccccccccccc} \toprule[1pt]
        \multirow{2}*{} & \multicolumn{2}{c}{Training}
        & \multicolumn{2}{c}{MSRA-B \cite{liu2011learning} }
        & \multicolumn{2}{c}{ECSSD \cite{yan2013hierarchical}}
        & \multicolumn{2}{c}{HKU-IS \cite{li2015visual}}
        & \multicolumn{2}{c}{PASCALS \cite{li2014secrets}}
        & \multicolumn{2}{c}{SOD \cite{movahedi2010design}}
         \\ \cmidrule(l){2-3} \cmidrule(l){4-5} \cmidrule(l){6-7} \cmidrule(l){8-9} \cmidrule(l){10-11} \cmidrule(l){12-13}
        Methods & Dataset & \#Images & \TheMeasures & \TheMeasures  & \TheMeasures
        & \TheMeasures & \TheMeasures  \\ \midrule[1pt]
        RC \cite{cheng2015global}& - & - &
        0.817 & 0.138 & 0.741 & 0.187 & 0.726 & 0.165 & 0.640 & 0.225 & 0.657 & 0.242 \\
        CHM \cite{li2013contextual} & - & - &
        0.809 & 0.138 & 0.722 & 0.195 & 0.728 & 0.158 & 0.631 & 0.222 & 0.655 & 0.249 \\
        DSR \cite{li2013saliency} & - & - &
        0.812 & 0.119 & 0.737 & 0.173 & 0.735 & 0.140 & 0.646 & 0.204 & 0.655 & 0.234 \\
        DRFI \cite{WangDRFI2017}& MB & 2,500 &
        0.855 & 0.119 & 0.787 & 0.166 & 0.783 & 0.143 & 0.679 & 0.221 & 0.712 & 0.215 \\
        MC \cite{zhao2015saliency}& MK & 8,000 &
        0.872 & 0.062 & 0.822 & 0.107 & 0.781 & 0.098 & 0.721 & 0.147 & 0.708 & 0.184 \\
        ELD \cite{lee2016deep}& MK & 9,000 &
        0.914 & 0.042 & 0.865 & 0.981 & 0.844 & 0.071 & 0.767 & 0.121 & 0.760 & 0.154 \\
        MDF \cite{li2015visual} & MB & 2,500 &
        0.885 & 0.104 & 0.833 & 0.108 & 0.860 & 0.129 & 0.764 & 0.145 & 0.785 & 0.155 \\
        DS \cite{li2016deepsaliency} & MB & 2,500 &
        - & - & 0.810 & 0.160 & - & - & 0.818 & 0.170 & 0.781 & 0.150 \\
        RFCN \cite{wangsaliency}& MK & 10,000 &
        0.926 & 0.062 & 0.898 & 0.097 & 0.895 & 0.079 & 0.827 & 0.118 & 0.805 & 0.161 \\
        DHS \cite{liu2016dhsnet}& MK + D & 9,500 &
        - & - & 0.905 & 0.061 & 0.892 & 0.052 & 0.820 & 0.091 & 0.823 & 0.127 \\
        $\text{DCL}^+$ \cite{li2016deep} & MB & 2,500 &
        0.916 & 0.047 & 0.898 & 0.071 & 0.907 & 0.048 & 0.822 & 0.108 & 0.832 & 0.126 \\ \midrule[1pt]
        Ours & MB & 2,500 &
        \textbf{0.927} & \textbf{0.028} & \textbf{0.915} & \textbf{0.052} & \textbf{0.913} & \textbf{0.039} & \textbf{0.830} & \textbf{0.080} & \textbf{0.842} & \textbf{0.118} \\
        $\text{Ours}^{\dagger} $ & MB & 2,500 &
        0.936 & 0.030 & 0.928 & 0.048 & 0.920 & 0.035 & 0.838 & 0.092 & 0.850 & 0.119 \\ \bottomrule[1pt]
    \end{tabular*}
    \caption{Quantitative comparisons with 11 methods on 5 popular datasets.
      The ResNet-101 \cite{He2016} version of our approach (\ie `Ours$\dagger$')
      clearly outperforms its VGGNet version.
      For fair comparison, we exclude `Ours$\dagger$' and
      highlight the best result of each column in \textbf{bold}.
      Here we use the initials of each dataset for convenience.
    }\label{tab:comparision}
\end{figure*}

\subsubsection{PR Curve}
We compare our approach with the existing methods in terms of PR curve here.
In \figref{fig:pr_curve},
we depict the PR curves produced by our approach and
previous state-of-the-art methods on 3 popular datasets.
It is obvious that FCN-based methods substantially outperform other methods.
More importantly, among all FCN-based methods,
the PR curve of our approach is especially outstanding
in the upper left corners of the coordinates.
We can also find that the precision of our approach
is much higher when the recall score is close to 1,
reflecting that our false positives are much lower than other methods.
This also indicates that our strategy of combining
low-level and high-level features in terms of short connections
is essential such that the resultant saliency maps
look much closer to the ground truth.

\subsubsection{F-measure and MAE}
We also compare our approach with the existing methods
in terms of F-meature and MAE scores.
The quantitative results are shown in \figref{tab:comparision}.
As can be seen, our approach achieves the best
score (maximum F-measure and MAE) on all datasets as listed in \figref{tab:comparision}.
On the ECSSD and SOD datasets,
our approach improves the current best maximum F-measure by 1 point,
which is a large margin as the values are already very close to ideal value 1.
In regard to MAE scores, our approach achieves a more than 1-point
decrease on MSRA-B and PASCALS datasets.
On the other datasets, there are still at least 0.09 points improvements.
This implies that the number of wrong predictions
in our case is significantly less than the other methods.

Besides, we also observe that the proposed
approach behaves even better on more difficult datasets,
such as HKUIS \cite{li2015visual}, PASCALS \cite{li2014secrets},
and SOD \cite{martin2001database, movahedi2010design}, which contain
a large number of images with multiple salient objects.
This indicates that our method is capable of detecting
and segmenting the most salient object,
while other methods often fail at one of these stages.

\subsection{The Existence of Saliency}

\begin{figure}
	\centering
    \small
	\setlength\tabcolsep{5pt}
	\begin{tabular}{cccc} \toprule[1pt]
		Methods & JSOD \cite{jiang2017joint} & MSRA-B \cite{liu2011learning} & ECSSD \cite{yan2013hierarchical} \\ \midrule[1pt]
        Wang~\etal \cite{wang2012salient} & 90.64\% & 89.26\% & 70.50\%  \\
		SSVM \cite{jiang2017joint} & \textbf{99.22\%} & 98.66\% & 94.40\%  \\
		Ours &  98.84\% &  \textbf{99.05\%} & \textbf{96.8\%}  \\ \bottomrule[1pt]
	\end{tabular}
	\caption{The prediction accuracy of our saliency existence branch compared to SSVM \cite{jiang2017joint}
    and Wang~\etal \cite{wang2012salient}. The best result of each column is highlighted in \textbf{bold}.}
    \label{fig:sal_exist}
    \vspace{-10pt}
\end{figure}

To date, most existing salient object detection methods
focus on datasets in which at least one salient object exists.
However, in many real-world scenarios, salient objects do not always exists.
Therefore, methods based on the above assumption may
easily lead to incorrect prediction results when applied to
scenes without any salient objects in them.
To solve this problem, we propose to introduce another
branch into our network to predict the saliency existence of the input image.
The new branch is composed of a global average pooling layer, followed
by a multi-layer perceptron (MLP) as the regressor to recognize the existence
of saliency as done in many classification networks \cite{simonyan2014very,He2016}.
The global average pooling layer is used to transform feature maps with different shapes into
the same size so that the resulting feature vectors can be fed into the MLP.
Like  \cite{simonyan2014very,girshick2015fast},
the MLP here consists of three fully-connected layers,
all of which are with 1,024 neurons except the last one which has two.
The softmax loss is used to optimize the new branch.

In our experiments, we use the same training set as in \cite{jiang2017joint},
which contains 5,000 background images (\ie images without salient objects in them)
and 5,000 images from MSRA10K \cite{cheng2015global}.
For these background images, the gradients from the salient
object detection module are not allowed to back-propagate
so that the resulting prediction maps would not be interfered.
We found that this operation is essential.
The hyper-parameters used here are the same to
our salient object detection experiments.
We train our network for 24,000 iterations and decrease the
learning rate by a factor of 10 at 20,000 iterations.
We test our model on three datasets, including JSOD \cite{jiang2017joint},
MSRA-B \cite{liu2011learning} and ECSSD \cite{yan2013hierarchical}.
\figref{fig:sal_exist} lists the results compared to another two works
SSVM \cite{jiang2017joint} and Wang~\etal \cite{wang2012salient}.
Since there is a clear separation between JSOD dataset
(mostly containing pure textures) and other two datasets
(MSRA-B and ECSSD mostly contain images with clear salient objects),
the classification results on all datasets have been already saturated
(very close to the ideal value ``1").
%
Thus, we expect more challenging dataset which better reflect
real world difficulties would be developed in near future.

\subsection{Timing}

Our network is fully convolutional,
which allows it to run very fast compared
against most previous salient object detection methods.
When trained on the MSRA-B dataset which contains 2,500 training images,
our network takes less than 8 hours for 12,000 iterations.
Interestingly, though 10,000 iterations are enough for convergence,
we found another 2,000 iterations still bring us a small performance gain in MAE.

During the inference stage, it takes us about 0.08s
to process an input image of size $300\times400$.
This is extremely faster than most of the previous works,
such as DCL \cite{li2016deep}
which need more than 1s for each image of the same size.
With our CRF layer considered, another 0.4 seconds are needed.
As a result, our overall time cost is less than 0.5s
for an image of size $300\times400$.

\renewcommand{\addFig}[1]{\includegraphics[width=0.158\linewidth]{Failure/#1}}
\renewcommand{\addFigs}[1]{\addFig{#1.jpg} & \addFig{#1.png} & \addFig{#1_dss.png}}

\begin{figure}[tp!]
	\centering
    \setlength\tabcolsep{1.2pt}
    \begin{tabular}{cccccc}
		\addFigs{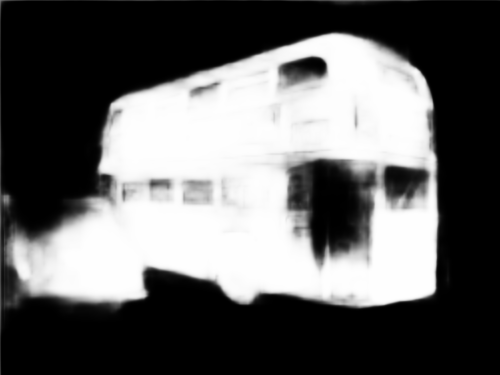} &
        \addFigs{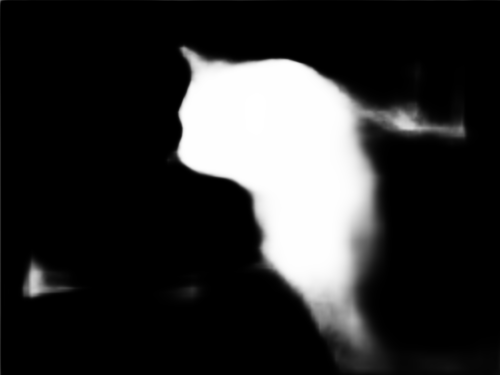} \\
		\addFigs{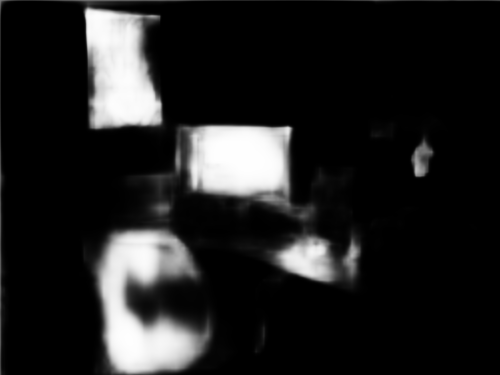} &
        \addFigs{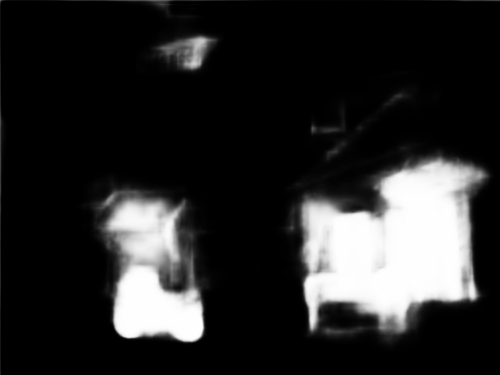} \\
        \addFigs{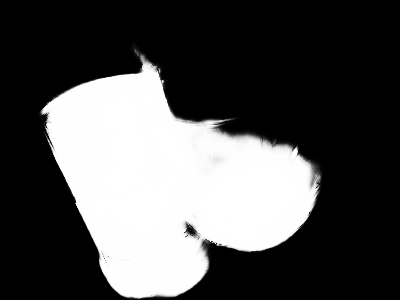} &
        \addFigs{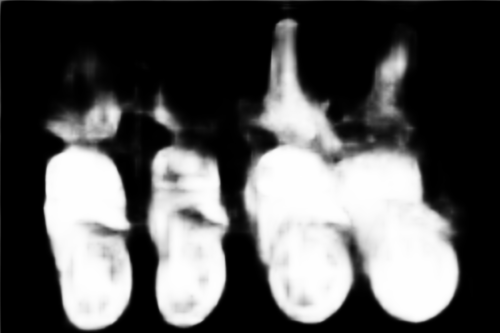} \\
        Source & GT & Ours & Source & GT & Ours
	\end{tabular}
	\caption{Failure cases selected from multiple datasets.
      As can be seen, most cases are caused by complex background,
      low contrast between foreground and background, and transparent objects.
    }\label{fig:failure_cases}
    \vspace{-10pt}
\end{figure}

\begin{figure*}
    \centering
    \small
    \setlength\tabcolsep{6.3pt}
    \begin{tabular*}{\textwidth}{lcccccccccccc} \toprule[1pt]
        \multirow{2}*{}
        & \multicolumn{2}{c}{MSRA-B  }
        & \multicolumn{2}{c}{ECSSD }
        & \multicolumn{2}{c}{HKU-IS }
        & \multicolumn{2}{c}{PASCALS }
        & \multicolumn{2}{c}{SOD }
        & \multicolumn{2}{c}{DUT-OMRON } \\
        \cmidrule(l){2-3}\cmidrule(l){4-5}\cmidrule(l){6-7}\cmidrule(l){8-9} \cmidrule(l){10-11} \cmidrule(l){12-13}
        Training Set & \TheMeasures & \TheMeasures  & \TheMeasures
        & \TheMeasures & \TheMeasures  & \TheMeasures \\ \midrule[1pt]
        MSRA-B (2500) & \textbf{0.920} & \textbf{0.043} & 0.908 & \textbf{0.064} & 0.902 & 0.049 & 0.824 & 0.101 & 0.836 & 0.126 & 0.764 & 0.070 \\
        ECSSD (1000) & 0.880 & 0.062 & - & - & 0.891 & 0.051 & 0.807 & 0.100 & \textbf{0.840} & \textbf{0.107} & 0.720 & 0.085 \\
        HKU-IS (2500) & 0.893 & 0.057 & 0.898 & 0.070 & \textbf{0.919} & \textbf{0.041} & 0.817 & \textbf{0.099} & 0.820 & 0.133 & 0.737 & 0.085 \\
        DUT-OMRON (3103) & 0.890 & 0.060 & 0.895 & 0.079 & 0.888 & 0.059 & 0.811 & 0.113 & 0.814 & 0.141 & \textbf{0.828} & \textbf{0.051} \\
        MSRA10K (6000) & - & - & \textbf{0.909} & 0.068 & 0.901 & 0.054 & \textbf{0.826} & 0.107 & 0.822 & 0.140 & 0.769 & 0.074 \\ \bottomrule[1pt]
    \end{tabular*}
    \caption{Performance when different training sets are used.
        The best results are highlighted in \textbf{bold}.
        Notice that all the results here are without CRF.
    }\label{tab:matrix_comparison}
    \vspace{-10pt}
\end{figure*}

\section{Discussion}

In this section, we conduct useful analysis on our proposed approach,
which we believe would be helpful for researchers to develop more powerful methods.

\subsection{Failure Case Analysis}

Some failure predictions of our approach have been shown
in \figref{fig:failure_cases}.
As can be seen, these failure cases can be categorized
into three circumstances in general.
The first one is actually the common defect of
CNN-based salient object detection methods,
in which the salient objects cannot be completely segmented out,
leaving a small part of the salient object missed.
Typical examples are the images shown in the first row
of \figref{fig:failure_cases}.
In the second circumstance, the main body of the salient object
cannot be extracted or non-salient regions are predicted to be salient.
As shown in the middle row of \figref{fig:failure_cases},
this case is mostly caused by complex backgrounds and very low contrast.
The last type of failure cases is caused by transparent objects
as shown in the bottom row of \figref{fig:failure_cases}.
Though our approach can detect some parts of the transparent objects,
to segment the complete objects out is still very difficult.

We argue that three possible remedies can be used to
solve the aforementioned problems.
First of all, a promising solution is to provide
more prior knowledge on segment level so that regions
with similar textures or colors can be detected simultaneously.
Because of the internal structure of CNNs,
the correlations of two positions in the score map
are decided by the learnable weights of the former layers,
making this problem difficult to be solved by the networks themselves.
Segment-level information allows CNNs to correct those
wrong predictions in the Circumstance 1 mentioned above.
In addition, segment-level information can also
serve as a post-processing tool to further refine
the predicted saliency maps by a simple voting strategy.
Secondly, more powerful training data should be presented,
including both simple and complex scenes.
As listed in \figref{tab:dataset_comparison},
training data with complex scenes can substantially
help improve the performance on both easy and difficult datasets.
Another solution should be designing more advanced models
and then extracting more powerful feature representations
to deal with challenging inputs with complex structures 
\cite{FanStructMeasureICCV17}.

\subsection{Benchmarking Training Set} \label{sec:train_set}

The selection of training set is one of the important
aspects for a learning based algorithm.
A good training set will definitely improve
the learning ability, leading to a more generative
model that can perform well on almost all scenes,
even with complex background.
%
%
However, the training sets of recent learning based approaches are
different and none of these works have explored which training set is the best.
\figref{tab:comparision} lists the details of different training sets
that existing approaches have used.
Furthermore, training on different datasets
with different sizes makes the comparisons unfair.
Albeit the number of training images is
not proportional to the performance gain,
the size and quality of different training sets
break the fair comparisons among different approaches.
One can observe in \figref{tab:comparision} that
some of them only use a training set with 2,500 images while
some others leverage around 10,000 images for training.

In this section, we attempt to thoroughly analyze the effect of
utilizing different datasets for training based on our proposed approach.
Our goal is to provide a new, unified, convincing,
and large-scale training set based on existing datasets for future research.
To do so, we perform a number of experiments and show exhaustive
comparisons among 6 widely-used and publicly available datasets,
which can be found in \figref{tab:matrix_comparison}.
Notice that all the training lists will be made \textit{publicly available}.
During testing phase, we use both the max F-measure score
and MAE score as measuring metrics.
Notice that since most datasets contain more than 5,000 images,
each model is trained for 16,000 iterations here.
An exception is the model trained on ECSSD with 6,000 iterations.

\subsubsection{Dataset Quality Measuring}

To exhibit the quality of datasets better,
each time we train on one of them, except for the SOD dataset
which has only 300 images and the PASCALS dataset
which has a lowly consistent behavior,
and test on all the test sets.
As ECSSD contains less than 2,000 images,
all the images are used for training and hence no image is left for testing.
For the remaining large-scale datasets,
if default splits are provided then they will be used directly.
Otherwise, we split the dataset in a ratio of
6:1:3 for training, validation, and testing, respectively.

Detailed experimental results have been shown in \figref{tab:matrix_comparison}.
As there is a large overlap between MSRA-B and MSRA10K datasets,
we only show the results on MSRA-B instead of both.
According to the results shown in \figref{tab:matrix_comparison},
the following conclusion can be drawn.
First, the best result on each dataset is always obtained
by training on the corresponding training set,
and the phenomenon is especially obvious for DUT-OMRON.
This might be caused by the characteristics of the images in each dataset,
making different datasets favor different features.
Consequently, we argue that it is \textit{inappropriate}
to directly compare performance numbers
that are achieved by different models trained
on different datasets (see also \figref{tab:matrix_comparison}).
%
Second, having more training images does not
necessarily entail better performance.
As can be seen in \figref{tab:matrix_comparison},
training on ECSSD dataset allows us to achieve the best performance
on the SOD dataset despite of having only 1,000 training images.
In regard to the above-mentioned issues,
a compromise solution is to construct a \textit{unified},
\textit{composite}, and \textit{versatile} dataset.

\begin{figure*}
	\centering
    \small
	\renewcommand{\tabcolsep}{6.9pt}
    \begin{tabular*}{\textwidth}{cccccccccccc} \toprule[1pt]
    	\multirow{2}*{Scheme} & \multirow{2}*{Training Set}
        & \multicolumn{2}{c}{MSRA-B}
        & \multicolumn{2}{c}{HKU-IS}
        & \multicolumn{2}{c}{PASCALS}
        & \multicolumn{2}{c}{SOD}
        & \multicolumn{2}{c}{DUT-OMRON}
        \\ \cmidrule(l){3-4} \cmidrule(l){5-6} \cmidrule(l){7-8} \cmidrule(l){9-10} \cmidrule(l){11-12}
         &  & \TheMeasures & \TheMeasures & \TheMeasures & \TheMeasures & \TheMeasures \\ \midrule[1pt]
        0 & MB (2500) & 0.920 & \textbf{0.043} & 0.902 & 0.049 & 0.824 & 0.101 & 0.836 & 0.126 & 0.764 & 0.070 \\
        1 & D + E (4103) & 0.901 & 0.053 & 0.907 & 0.048 & 0.832 & 0.090 & 0.846 & 0.109 & 0.832 & 0.050 \\
        2 & H + E (3500) & 0.897 & 0.054 & 0.923 & \textbf{0.040} & 0.825 & 0.092 & 0.849 & \textbf{0.108} & 0.753 & 0.078 \\
        3 & D + H (5603) & 0.905 & 0.053 & 0.924 & 0.042 & 0.832 & 0.096 & 0.839 & 0.130 & 0.833 & 0.052 \\
        4 & MB + E (3500) & 0.916 & 0.045 & 0.909 & 0.045 & 0.835 & 0.091 & 0.852 & 0.111 & 0.758 & 0.073 \\
        5 & MB + H (5000) & 0.920 & 0.045 & 0.925 & \textbf{0.040} & 0.834 & 0.095 & 0.845 & 0.121 & 0.774 & 0.072 \\
        6 & MB + D (5603) & 0.921 & 0.046 & 0.910 & 0.050 & 0.837 & 0.099 & 0.845 & 0.127 & 0.840 & \textbf{0.049} \\
        7 & MB + E + D (6603) & 0.921 & 0.046 & 0.915 & 0.048 & 0.842 & 0.091 & 0.858 & 0.115 & 0.839 & 0.051 \\
        8 & MB + H + D (8103) & \textbf{0.923} & 0.046 & 0.926 & 0.043 & 0.837 & 0.096 & 0.855 & 0.123 & 0.840 & 0.051 \\
        9 & MB + E + H (6000) & 0.921 & 0.045 & 0.926 & \textbf{0.040} & 0.841 & 0.090 & 0.860 & 0.111 & 0.786 & 0.069 \\
        10 & E + D + H (6603) & 0.911 & 0.050 & 0.925 & 0.041 & \textbf{0.844} & \textbf{0.087} & 0.854 & 0.110 & 0.835 & 0.051 \\
        11 & MB + E + D + H (9103) & \textbf{0.923} & 0.046 & \textbf{0.927} & 0.042 & \textbf{0.844} & 0.091 & \textbf{0.864} & 0.113 & \textbf{0.843} & 0.051 \\ \bottomrule[1pt]
    \end{tabular*}
    \caption{Detailed information of different training sets
      and the corresponding results on 5 datasets.
      The best results are highlighted in {\em bold}.
      All the results are obtained without any post-processing.
      Here we use the initials of each dataset for convenience.
    }\label{tab:dataset_comparison}
\end{figure*}

\subsubsection{Beyond Training on Individual Datasets}

We select 4 datasets from \figref{tab:matrix_comparison}
to build composite datasets for comparisons.
Though the MSRA10K is more than twice bigger than MSRA-B dataset,
models trained on it have a competitive performance
compared to those trained on the MSRA-B dataset.
Here we just keep MSRA-B for training due to its
high-quality images and annotations.
Therefore, there are totally 11 different combinations
which have been shown in the second column of \figref{tab:dataset_comparison}.
During the testing phase,
we also use the six test sets mentioned above for fair comparisons.

From the results in \figref{tab:dataset_comparison},
the following conclusions can be drawn.
First of all, a larger training set does not necessarily
mean higher test performance.
This phenomenon can be observed through comparing
Scheme 3 with other schemes.
Despite only 3,500 training images,
this combination performs better than those with more than 6,000 training images.
It is true that the quality of annotations
might be an essential reason that causes such a problem.
However, such a consideration is beyond the scope of this paper.
All conclusions here are based on the assumption
that each dataset we use is with well-segmented annotations.

%
%
%

Second, an inappropriate combination of datasets may
result in worse performance compared with individual datasets.
By comparing schemes 4 and 0, one can find that despite
better performance on HKU-IS, PASCALS, and SOD datasets
there are still slight decreases when testing on MSRA-B and DUT-OMRON datasets.
%


Through this series of experiments, we aimed to emphasis that
a training set with a large quantity of images may not
be capable of bringing in better performance gain.
A good training set should take into account as many cases as possible.
However, because of the diversity of existing datasets,
it is hard to obtain a convincing dataset that
can behave the consistency among all existing datasets.
In regard to the current state in salient object detection,
we recommend using our Scheme 11 in \figref{tab:dataset_comparison}
as training set for fair comparison and fitting decreasing
performance bias caused by different training sets.
Another severe problem in salient object detection is
that most datasets are no longer challenging.
An explicit effect is that the differences between different models
are difficult to be distinguished because of
the close performance on existing datasets.
We hope that more challenging datasets with complex scenes
and high consistency would be presented in the near future.


\section{Conclusion}


In this paper, we presented a deeply supervised network
for salient object detection.
Instead of directly connecting loss layers to the last
layer of each stage, we introduce a series of short
connections between shallower and deeper side-output
layers.
With these short connections, the activation of each
side-output layer gains the capability of both
highlighting the entire salient object and accurately locating
its boundary.
A fully connected CRF is also employed for correcting
wrong predictions and further improving spatial coherence.
Our experiments demonstrate that these mechanisms result
in more accurate saliency maps over a variety of images.
Our approach significantly advances the state-of-the-art and is capable of
capturing salient regions in both simple and difficult cases,
which further verifies the merit of the proposed architecture.

\section*{Acknowledgments}
This research was supported by NSFC (NO. 61620106008, 61572264),
Huawei Innovation Research Program,
CAST YESS Program,
and IBM Global SUR award.

\if CLASSOPTIONcaptionsoff
\newpage
\fi

\bibliographystyle{IEEEtran}
\bibliography{DeepSal}

\vspace{-.3in}
\begin{IEEEbiography}[{\includegraphics[width=1in,keepaspectratio]{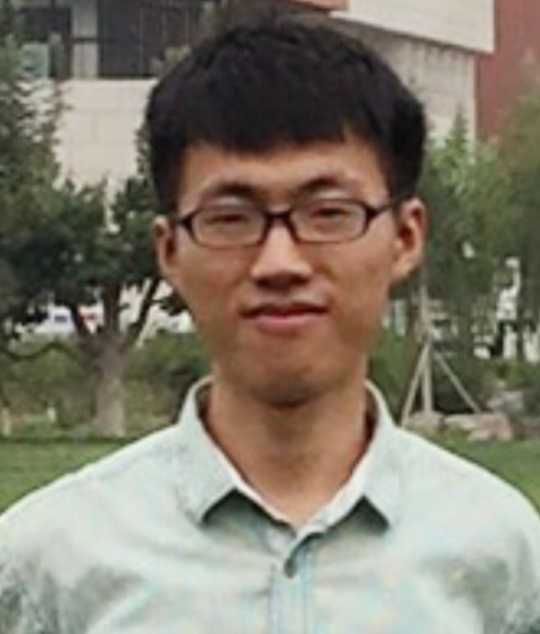}}]
{Qibin Hou} is currently a Ph.D. Candidate with College of Computer Science and
Control Engineering, Nankai University, under the supervision of Prof. Ming-Ming Cheng.
His research interests include deep learning, image processing, and computer vision.
\end{IEEEbiography}

\vspace{-.5in}
\begin{IEEEbiography}[{\includegraphics[width=1in,keepaspectratio]{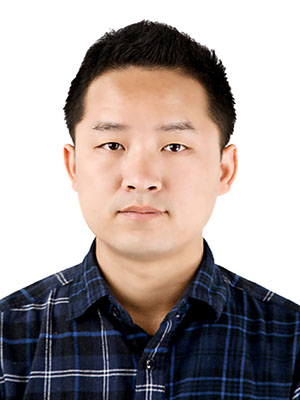}}]
{Ming-Ming Cheng} received his PhD degree from Tsinghua University in 2012.
Then he did 2 years research fellow, with Prof. Philip Torr in Oxford.
He is now a professor at Nankai University, leading the Media Computing Lab.
His research interests includes computer graphics, computer vision, and image processing.
He received research awards including ACM China Rising Star Award,
IBM Global SUR Award,
CCF-Intel Young Faculty Researcher Program, \etc
\end{IEEEbiography}

\vspace{-.4in}
\begin{IEEEbiography}[{\includegraphics[width=1in,keepaspectratio]{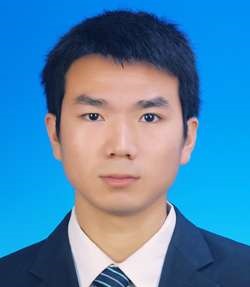}}]
{Xiaowei Hu} is currently a Master student
with College of Computer Science and
Control Engineering, Nankai University,
under the supervision of Prof. Ming-Ming Cheng.
His research interests include deep learning,
image processing, and computer vision.
\end{IEEEbiography}

\vspace{-.5in}
\begin{IEEEbiography}[{\includegraphics[width=1in,keepaspectratio]{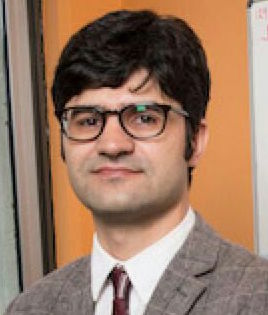}}]
{Ali Borji}
received the PhD degree in cognitive
neurosciences from the Institute for Studies in
Fundamental Sciences (IPM), 2009.
He is currently an assistant professor at
Center for Research in Computer Vision,
University of Central Florida.
His research interests include visual
attention, visual search, machine learning, neurosciences,
and biologically plausible vision models.
\end{IEEEbiography}

\vspace{-.4in}
\begin{IEEEbiography}[{\includegraphics[width=1in,keepaspectratio]{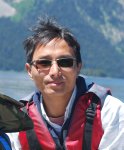}}]
{Zhuowen Tu} received the BE degree from the
Beijing Information Technology Institute,
the ME degree from Tsinghua University,
and the PhD degree in computer science from Ohio State University.
He is an associate professor of cognitive science
with the University of California, San Diego (UCSD).
His main research interests include computer vision,
machine learning, and neural computation.
\end{IEEEbiography}

\vspace{-.4in}
\begin{IEEEbiography}[{\includegraphics[width=1in,keepaspectratio]{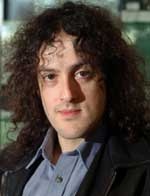}}]
{Philip H.S. Torr} received the PhD degree from Oxford University.
After working for another 3 years at Oxford,
he worked for 6 years as a research scientist for Microsoft Research,
first in Redmond, then in Cambridge,
founding the vision side of the Machine Learning and Perception Group.
He is now a professor at Oxford University.
He has won awards from several top vision conferences,
including ICCV, CVPR, ECCV, NIPS \etc
He is a Royal Society Wolfson Research Merit Award holder.
\end{IEEEbiography}
	
\vfill

\end{document}